*Article*

# An Analysis of Layer-Freezing Strategies for Enhanced Transfer Learning in YOLO Architectures

Andrzej D. Dobrzycki *, Ana M. Bernardos and José R. Casar

Information Processing and Telecommunications Center, Universidad Politécnica de Madrid,
ETSI Telecomunicación, Av. Complutense, 30, 28040 Madrid, Spain; anamaria.bernardos@upm.es (A.M.B.);
joseramon.casar@upm.es (J.R.C.)
* Correspondence: daniel.dobrzycki@upm.es

**Abstract**

The You Only Look Once (YOLO) architecture is crucial for real-time object detection. However, deploying it in resource-constrained environments such as unmanned aerial vehicles (UAVs) requires efficient transfer learning. Although layer freezing is a common technique, the specific impact of various freezing configurations on contemporary YOLOv8 and YOLOv10 architectures remains unexplored, particularly with regard to the interplay between freezing depth, dataset characteristics, and training dynamics. This research addresses this gap by presenting a detailed analysis of layer-freezing strategies. We systematically investigate multiple freezing configurations across YOLOv8 and YOLOv10 variants using four challenging datasets that represent critical infrastructure monitoring. Our methodology integrates a gradient behavior analysis (L2 norm) and visual explanations (Grad-CAM) to provide deeper insights into training dynamics under different freezing strategies. Our results reveal that there is no universal optimal freezing strategy but, rather, one that depends on the properties of the data. For example, freezing the backbone is effective for preserving general-purpose features, while a shallower freeze is better suited to handling extreme class imbalance. These configurations reduce graphics processing unit (GPU) memory consumption by up to 28% compared to full fine-tuning and, in some cases, achieve mean average precision (mAP@50) scores that surpass those of full fine-tuning. Gradient analysis corroborates these findings, showing distinct convergence patterns for moderately frozen models. Ultimately, this work provides empirical findings and practical guidelines for selecting freezing strategies. It offers a practical, evidence-based approach to balanced transfer learning for object detection in scenarios with limited resources.

**Key Contribution:** This manuscript presents a comprehensive layer-freezing analysis for YOLOv8 and YOLOv10 models, demonstrating up to a 28% reduction in GPU usage while maintaining or improving accuracy compared to full fine-tuning across four infrastructure-monitoring datasets (InsPLAD-det, Electric Substation, Common-VALID, and Bird's Nest). We identify optimal freezing strategies, such as freezing the first four blocks or the complete backbone, that balance efficiency and performance, supported by in-depth analysis of training dynamics including gradient L2 norm evolution and GradCAM activation patterns.

## 1. Introduction

The You Only Look Once (YOLO) architecture has revolutionized real-time object detection by enabling detection, localization, and classification in a single forward pass, making it particularly suitable for time-sensitive applications such as surveillance, autonomous driving, and UAV-based monitoring [1–4]. However, balancing high detection accuracy with computational demands remains challenging, especially for deployment in resource-constrained environments like edge devices and embedded systems.

Transfer learning through layer freezing has emerged as a promising strategy to address these constraints by adapting pre-trained models to new tasks without complete retraining, thereby preserving learned knowledge while reducing computational overhead [5,6]. While existing research demonstrates that transfer learning can enhance model performance when adapting models pre-trained on large-scale datasets like COCO [7] to domain-specific datasets [8,9], a significant gap exists in the systematic analysis of layer freezing strategies within modern YOLO architectures.

Specifically, there is insufficient comprehensive investigation into how different layer freezing configurations impact the trade-offs between key performance metrics (mAP@50, mAP@50:95), training efficiency, and resource utilization in YOLOv8 and YOLOv10 architectures. This understanding is crucial for practitioners optimizing YOLO models for real-world applications where both accuracy and efficiency are paramount [10,11].

This research systematically investigates the impact of various layer-freezing strategies on contemporary YOLO architectures (YOLOv8 [12] and YOLOv10 [13]) applied to critical infrastructure monitoring tasks. We evaluate different freezing configurations against traditional fine-tuning across four challenging datasets, analyzing accuracy metrics, computational efficiency, and training dynamics.

The remainder of this paper is organized as follows: Section 2 reviews related work, Section 3 describes the YOLO architectures, Section 4 details the datasets, Section 5 outlines the experimental methodology, Section 6 presents the results, Section 7 discusses findings and limitations, and Section 8 concludes the work.

## 2. Related Work

The concept of layer freezing in deep neural networks has emerged as a crucial technique for optimizing transfer learning, particularly in scenarios where computational efficiency is paramount. Transfer learning leverages models that are pre-trained on large datasets to adapt to new, often smaller, task-specific datasets through fine-tuning, which adjusts all parameters, or selective freezing, which fixes certain layers while allowing others to adapt [5]. Layer freezing not only accelerates convergence but also reduces memory consumption and communication costs in distributed training, and it helps mitigate overfitting when target datasets are limited [10,11,14].

Several strategies have been proposed for optimizing layer freezing. The work in [10] introduced AutoFreeze, a dynamic approach that determines which layers to freeze based on training progress, demonstrating effectiveness in both natural language processing (NLP) and computer vision (CV) tasks. Similarly, prior works have explored layer freezing in transformer architectures, showing substantial improvements in training efficiency with minimal performance trade-offs [11,14,15]. Hybrid approaches combining freezing with pruning have also shown promise, particularly in resource-constrained environments [16,17].

In object detection, however, the systematic analysis of layer freezing remains limited despite its potential for optimizing computationally demanding architectures like YOLO. Early work by [8] investigated freezing backbone layers in YOLOv5 models pre-trained on COCO and fine-tuned on smaller datasets, finding that frozen backbones achieved higher mAP@50 scores than fully fine-tuned models. The study in [9] extended this analysis to vehicle detection, confirming superior performance with frozen backbones, particularly for imbalanced classes. Recent work by [18] extended this analysis to UAV-based object detection, showing that backbone freezing combined with architectural modifications can improve performance while reducing computational overhead. However, the research in [19] noted limitations in fine-grained detection tasks, where frozen models missed subtle details despite faster inference.

The emergence of modern YOLO architectures has prompted more sophisticated transfer learning approaches. The work in [20] systematically evaluated three transfer learning methods, frozen layers, domain adaptation, and zero-shot learning, in YOLOv8 for architectural drawings, finding that selective layer freezing (particularly intermediate layers between neck and head) achieved the highest mAP. Most recently, the study in [21] demonstrated that deeper fine-tuning in YOLOv8n yields substantial performance gains on fine-grained tasks while maintaining negligible performance degradation on the original COCO benchmark.

Table 1 summarizes key findings from recent studies. Despite progress, modern YOLO architectures (YOLOv8, YOLOv10) lack comprehensive evaluation across real-world scenarios, focusing primarily on specific applications or single freezing strategies without systematic evaluation of gradient dynamics, computational trade-offs, or performance across varied dataset characteristics. This work fills that gap through a detailed analysis of freezing strategies for critical infrastructure monitoring, balancing accuracy and efficiency.

**Table 1.** Summary of key studies on layer freezing in object detection.

| Study | Architecture | Techniques Used | Key Findings | Limitations |
|---|---|---|---|---|
| [20] | YOLOv8 | Frozen layers, domain adaptation, zero-shot learning | Freezing neck-head layers achieved highest mAP (0.707) | Domain adaptation needs refinement for specific features |
| [21] | YOLOv8 Nano | Progressive backbone unfreezing, dual-head evaluation | Deep fine-tuning (+10% mAP) with minimal forgetting (<0.1%) | Limited to a single fruit detection dataset; limited to a single model variant |
| [18] | YOLOv5 Large | Backbone freezing, concatenation modification, multi-head detection | Combined modifications achieved 22.4% mAP | Frozen backbone alone reduced performance significantly; limited to VisDrone-DET2019 dataset; limited to a single model variant |
| [15] | ViTDet | Adaptive parameter freezing (gradient-based), linear freezing | 1.40× speedup with 2.5% max accuracy loss | Understanding of convergence dynamics needs improvement |
| [8] | YOLOv5 | Backbone freezing | Superior performance on small/imbalanced datasets | Limited to single architecture evaluation; limited to PWMFD dataset |
| [19] | YOLOv5 | Backbone freezing | Frozen models are faster but miss fine-grained details | Trade-off between speed and precision |
| [9] | YOLOv5 | Backbone freezing | Improved vehicle detection with frozen features | Task-specific validation only; limited to a single vehicle detection dataset |

## 3. Background: You Only Look Once Architectures

In our research, we have chosen to work with the YOLOv8 [12] and YOLOv10 [13] architectures due to their significant advancements and optimizations over their predecessors, as well as due to their state-of-the-art performance and efficiency. These architectures have been developed to improve the performance, accuracy, and efficiency of object detection tasks.

*3.1. YOLOv8*

YOLOv8, primarily a series of improvements and enhancements made to the YOLOv5 architecture, represents a significant evolution in the YOLO family. It incorporates several innovative features that enhance its performance in various object recognition tasks. While many of the changes in YOLOv8 are related to model scaling and architecture optimizations, addressing some of the limitations seen in previous versions, this research will provide insights into the changes most relevant to the layer freezing of the network:

- Enhanced feature pyramid network (FPN): The YOLOv8 algorithm employs an advanced version of the FPN, designated as Spatial Pyramid Pooling—Fast (SPPF), which markedly enhances its capacity to identify objects across a spectrum of scales. This enhancement is particularly crucial for detecting small objects in images, a common challenge in many real-world applications.
- Improved backbone network: The backbone network in YOLOv8 has been significantly optimized for better feature extraction. It uses a similar backbone as YOLOv5 with modifications to the CSPLayer, now called the C2f module (cross-stage partial bottleneck with two convolutions). The C2f module combines high-level features with contextual information, enhancing detection accuracy. Additionally, the backbone employs CSPNet (Cross Stage Partial Network) [22], which helps in reducing the computational cost while maintaining high accuracy. The backbone is specifically the CSPDarknet53 [23] feature extractor, known for its efficiency and effectiveness in extracting features.

These enhancements to the FPN and backbone network underscore their critical role in the overall YOLOv8 architecture. The architecture consists of three main components: the backbone, the neck, and the head. The backbone, which is responsible for feature extraction, is typically the most important part of the model to maintain during transfer learning. In the case of YOLOv8, the backbone consists of the first 9 blocks of the model [24], each block with a variable number of layers, being the SPPF block part of the neck. By focusing on this key component, we aim to ensure that the basic feature extraction capabilities are preserved, which is critical to the success of the transfer learning process. Figure 1 provides an overview of the key components of the YOLOv8 architecture relevant to our experiment, highlighting the blocks we will be manipulating when applying freezing techniques.

A deeper analysis of the YOLOv8 configuration shows that blocks 0–3 are dedicated to pure feature extraction layers, strategically placed ahead of any feature concatenation processes. Specifically, these initial blocks include the following: (i) block 0 with initial convolution and P1/2 downsampling, (ii) block 1 with P2/4 downsampling convolution, (iii) block 2 featuring the C2f module for feature processing, and (iv) block 3 implementing P3/8 downsampling. The first feature concatenation occurs at block 14 in the head, where backbone features from block 4 (P3/8 level) are concatenated with upsampled features via concatenation. This architectural design creates a natural boundary where blocks 0–3 capture fundamental, domain-agnostic visual representations (edges, textures, basic geometric patterns), while subsequent blocks engage in task-specific feature fusion operations. The backbone encompasses blocks 0–8, with the SPPF module at block 9 marking the transition to the neck component.

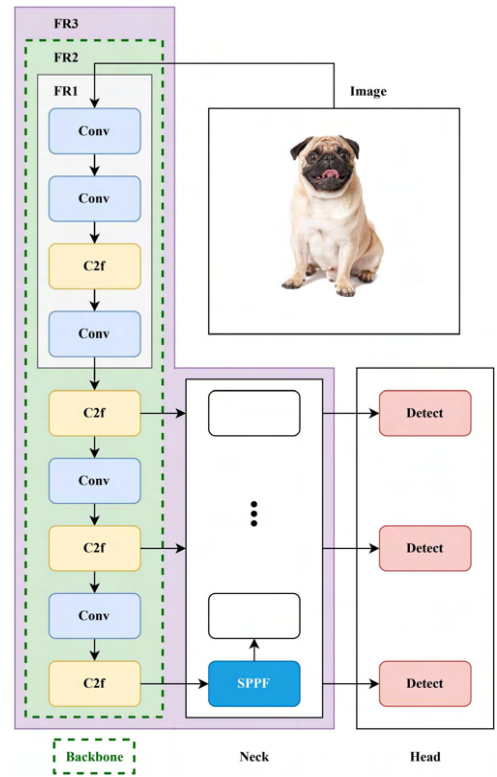

**Figure 1.** Simplified YOLOv8 architecture highlighting the layer-freezing strategy implementation. The model consists of backbone (feature extraction), neck (feature fusion), and head (detection) components. Highlighted regions FR1, FR2, and FR3 indicate the progressive freezing approaches: FR1 freezes the first 4 blocks (1% of parameters), FR2 freezes the complete backbone encompassing 9 blocks (35–44% of parameters), and FR3 freezes 22 blocks, including most of the neck (72–87% of parameters). This architectural overview demonstrates how different freezing strategies preserve varying levels of pre-trained knowledge while allowing targeted adaptation.

*3.2. YOLOv10*

By August 2024, YOLOv10 became the latest evolution of the YOLO series, representing the current state of the art in real-time, end-to-end object detection [13]. It introduces several significant improvements over its predecessors, particularly YOLOv8. In line with our primary objective of investigating the impact of layer freezing on training efficiency and accuracy metrics, the main enhancements incorporated into YOLOv10 are as follows.

- Enhanced model architecture: YOLOv10 introduces several architectural innovations, such as the lightweight classification head, spatial-channel decoupled downsampling, and rank-guided block design. These changes are aimed at reducing computational redundancy, improving parameter utilization, and enhancing overall model efficiency. Furthermore, YOLOv10 integrates a partial self-attention module to boost model capability, thereby achieving better accuracy with minimal computational overhead.
- Transformer block integration: YOLOv10 incorporates transformer blocks, specifically multi-head self-attention (MHSA) followed by feed-forward networks (FFNs). This integration helps in capturing global contextual information more effectively, leading to improved detection accuracy. By placing the partial self-attention (PSA) module only after the last stage, the model minimizes the computational complexity while enhancing global representation learning capabilities.

YOLOv10 maintains the fundamental architectural principle established in YOLOv8 while incorporating advanced attention mechanisms. The initial four-block boundary (blocks 0–3) remains consistent, preserving the same pure feature extraction pattern: initial convolutions with P1/2 and P2/4 downsampling, C2f feature processing, and P3/8 downsampling. Critically, YOLOv10's architectural enhancements, including the partial self-attention (PSA) modules and transformer blocks with multi-head self-attention (MHSA) and feed-forward networks (FFNs) are strategically positioned after this initial feature extraction phase. The PSA modules are integrated primarily in the later stages of the backbone and neck, allowing the fundamental convolutional feature extraction in blocks 0–3 to remain unaltered. The rank-guided block design optimizes parameter utilization while maintaining the architectural boundaries that are crucial for transfer learning. This design ensures that the domain-agnostic feature extraction capabilities are preserved in the initial blocks, while the enhanced attention mechanisms provide task-specific adaptation capabilities in subsequent processing stages.

These advancements highlight the intricate architecture of YOLOv10, which consists of a backbone, a neck, and a head. The backbone of YOLOv10 was designed following a rank-guided block style to maximize feature extraction while maintaining computational efficiency. Specifically, the backbone in YOLOv10 includes 9 blocks, optimized to extract deep and rich features essential for high-accuracy detection tasks. Additionally, the SPPF and PSA blocks, assumed to be part of the neck following the style presented in [24], contribute to the overall architecture. Figure 2 presents a simplified view of the main components of the YOLOv10 architecture pertinent to our experiment, focusing on the blocks we will adjust during the freezing process for clearer comprehension.

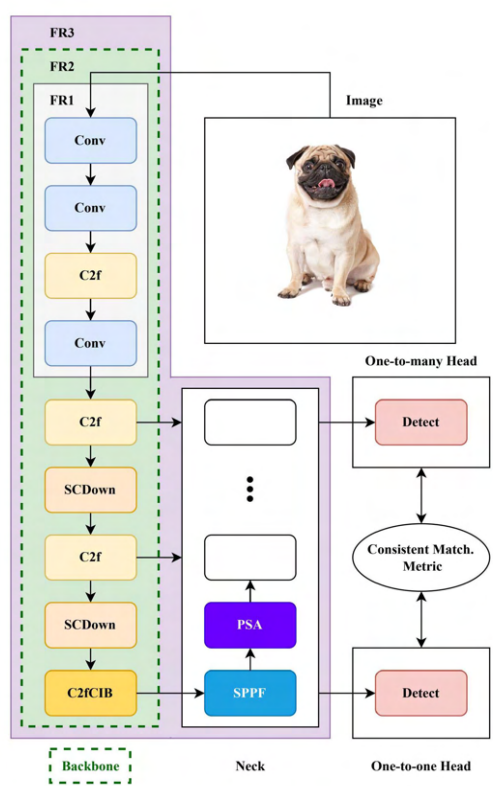

**Figure 2.** Simplified YOLOv10 architecture with layer-freezing regions for transfer learning optimization. The enhanced architecture incorporates transformer blocks and attention mechanisms while maintaining the three-component structure. Freezing regions FR1, FR2, and FR3 represent escalating parameter preservation strategies: FR1 targets initial feature extraction layers (1–2% of parameters), FR2 encompasses the complete backbone including attention modules (28–51% of parameters), and FR3 extends to 23 blocks covering most processing stages (67–89% of parameters). The architecture demonstrates the integration of modern attention mechanisms within the freezing framework.

## 4. Application & Datasets

This section presents the applications and datasets used, emphasizing object detection for critical infrastructure inspection and surveillance. UAVs play a key role in monitoring remote or hazardous environments such as power lines and industrial sites. Real-time object detection enhances both safety and operational efficiency in these scenarios.

To rigorously evaluate the models, we selected four datasets that mirror the challenges of these real-world scenarios, providing diverse environments to thoroughly test the models' adaptability and precision. We incorporate both real-world and synthetic datasets to reflect current trends in leveraging synthetic data for training machine learning models, as synthetic datasets enable the creation of controlled environments that simulate rare or hazardous conditions. This comprehensive coverage ensures that our layer freezing analysis is robust across diverse operational scenarios and technical requirements. The datasets represent a carefully curated combination of real-world and synthetic data, addressing different aspects of the infrastructure monitoring domain:

- Domain diversity: coverage spans from detailed substation equipment inspection to transmission line asset monitoring and general aerial surveillance scenarios
- Technical challenges: each dataset presents unique computational and detection challenges, from multi-scale object detection to extreme class imbalance
- Environmental variability: real-world datasets incorporate varying weather conditions, lighting scenarios, and seasonal changes
- Data acquisition methods: multiple UAV platforms, camera systems, and imaging perspectives are represented

*4.1. InsPLAD Dataset*

The InsPLAD [25] dataset is specifically designed for power line asset inspection, created from UAV images captured during real-world inspections. This dataset fills a critical gap in the literature by providing comprehensive coverage of both asset detection and defect classification tasks. Our work focuses on the InsPLAD-det portion of this dataset, which is used for object detection.

InsPLAD-det: This portion of the dataset includes 17 different object classes, with a total of 28,933 annotations across 10,607 high-resolution RGB images (1920 × 1080 pixels) captured using a DJI Matrice 210 V2 drone equipped with a Zenmuse z30 camera (DJI, Shenzhen, China). Images were systematically captured between October and December 2020 from 226 transmission towers in Brazil's 500 kV high-voltage network. A standardized capture protocol ensured consistent image quality, with key components centered in frames and specific angles maintained to enhance inspection efficacy. The collection process incorporated multiple environmental conditions, flight altitudes, and perspectives to simulate real-world operational scenarios.

The distribution of objects and annotations covers a wide range of transmission line assets, including insulators, conductors, towers, and associated hardware components. The assets vary widely in terms of the number of instances and their spatial properties. For example, the class with the fewest images is "Polymer Insulator Tower Shackle", with 57 images and 57 annotations, while the class with the most images is "Polymer Insulator", with 3173 images and 3244 annotations.

The dataset is split with 80% of the images used for training and 20% for testing and validation. After retrieving the dataset from its original source, and accounting for some missing annotated images, a total of 10,533 images with 28,869 annotations were gathered. Following the data partitioning, the dataset comprised 7907 images for training and 2626 images for testing/validation.

*4.2. Electric Substation Dataset*

The Electric Substation [26] dataset provides an extensively annotated resource for substation equipment inspection, addressing the automation needs of electrical distribution infrastructure monitoring. It consists of 7539 images of electric substations with 213,566 annotated objects, captured using various cameras, including cellphones, panoramic aerial cameras, stereo FLIR cameras, and automated guided vehicles (AGVs). Collected over two years from a single electrical distribution substation in Brazil, the images represent different times of day, weather, and seasonal conditions.

The dataset is organized into several directories based on the time of collection and the device used. It includes 4030 images collected manually during the morning period (8:00 to 12:00). There are 2270 images captured using an AGV during daytime hours, both in the morning (8:00 to 10:00) and in the afternoon (13:00 to 17:00). Additionally, the dataset contains 899 images collected using an AGV during nighttime with artificial lighting (20:00 to 21:00) and 340 images taken using an AGV during nighttime without artificial light sources (20:00 to 21:00).

It includes 15 object classes with severe class imbalances. After a small percentage of unlabeled images are removed, the class with the fewest images is "Open blade disconnect switch", with 591 images and 1076 annotations, while the class with the most images is "Porcelain Pin Insulator", with 6457 images and 107,497 annotations. This extreme imbalance reflects real-world operational conditions under which certain equipment types are significantly more prevalent.

For more details about the distribution between class images and instances on the Electric Substation dataset, please refer to Appendix B.1. Since the Electric Substation dataset was not originally partitioned for training, testing, and validation, we applied a 70-20-10 split, ensuring that class proportions were maintained across all subsets. This resulted in a total of 7076 images and 202,509 object annotations. After the division, the dataset consisted of 5023 images for training, 1452 images for testing, and 601 images for validation.

*4.3. Common-VALID Dataset*

The Common-VALID dataset, derived from the comprehensive VALID dataset [27], represents the integration of synthetic data generation in infrastructure monitoring research. This synthetic dataset addresses the limitations and costs associated with real-world data collection while providing controlled environmental conditions for systematic evaluation.

The parent VALID dataset consists of 6690 high-resolution images, all annotated with panoptic segmentation across 30 categories, traditional and oriented bounding boxes for object detection, and binocular depth maps. Images were captured across six virtual scenes and five environmental conditions (sunny, sunset, night, snow, fog) at multiple altitudes (20 m, 50 m, 100 m). It includes 30 categories, 17 of them for object detection, categorized into "thing" and "stuff" types. "Thing" categories encompass objects like trees, buildings, and vehicles, while "stuff" categories include land, water, and ice. This hierarchical categorization aids in detailed and accurate object detection and segmentation tasks.

We curated the VALID dataset by selecting the most representative classes for common transfer learning scenarios. This streamlined dataset allowed us to evaluate the effects of freezing layers in various YOLO architectures, comparing them to training from scratch or fine-tuning the models. The selected eight classes (small vehicle, large vehicle, animal, person, ship, plane, bridge, harbor) represent common objects encountered in aerial surveillance and infrastructure monitoring applications. The dataset presents significant object size variation, with 25% of instances below 10 pixels, creating ultra-small object detection challenges. This distribution reflects real-world aerial imaging scenarios where objects appear at various scales depending on altitude and camera settings.

The creation process involved filtering the VALID dataset to retain only the classes most relevant for our experiments, ensuring a streamlined dataset for evaluating the impact of freezing layers, adjusting bounding box labels for accuracy and relevance, and splitting the dataset into training, test, and validation subsets. Training, test, and validation sets were created according to a ratio of 70-15-15, ensuring that the class proportions are preserved in each subset. After partitioning, the Common-VALID dataset contains 6054 images and 70,799 annotations, distributed as 4246 images for training, 908 for testing, and 900 for validation. The class with the fewest images is "Plane", consisting of 268 images and 429 annotations, while the class with the most images is "Small vehicle", which includes 4724 images and 40,230 annotations. For further details on the class distribution and annotations within the Common-VALID dataset, please refer to Appendix B.2.

*4.4. Bird's Nest Dataset*

The Bird's Nest [28] dataset is tailored to the detection of nests on power transmission towers, based on aerial photos taken via UAVs, and thus, it presents a specialized single-class detection challenge.

Originally comprising 800 aerial images, our accessible subset contains 401 images captured via UAV platforms. The dataset poses unique challenges for single-class detection in aerial scenarios. First, the high-altitude perspective introduces scale and perspective distortions that complicate object localization. Second, natural occlusions caused by vegetation, tower structures, and environmental elements create complex visual obstructions. Lastly, the texture similarity between nest materials and their surroundings, such as natural backgrounds and structural elements, makes visual discrimination difficult.

To prepare it for our experiments, we split the dataset into training, testing, and validation subsets with a 70-20-10 ratio, resulting in 281 images for training, 80 for testing, and 41 for validation. To mitigate the limitations of the training dataset size, we employed a range of data augmentation techniques, following the implementation described in [28]. These included horizontal and vertical flips, rotations, random cropping, Gaussian blur, and the addition of Gaussian noise. These augmentations were essential to simulate different real-world conditions and increase the diversity of the training data. Through these techniques, we expanded the training set to a total of 3359 images. Additionally, we adjusted the validation set by adding 359 augmented images from the training subset, resulting in a final distribution of 3000 training images, 400 validation images, and 80 testing images. Almost all images contain only a single instance per image, with 3782 annotations in total.

The datasets used in this article are comprehensive resources for training and evaluating object detection models, presenting challenges such as class imbalance and diverse conditions. They support the development of computer vision and deep learning algorithms for automating inspection and maintenance in various fields. The inclusion of synthetic data, which is used in recent studies [29–31], offers potential advantages such as increased data diversity and the ability to simulate rare or dangerous conditions without the need for real-world data collection. These datasets make a consistent benchmark for our research on layer freezing in YOLO architectures and are valuable assets for researchers and engineers aiming to advance automated inspection technologies.

Table 2 summarizes how each dataset addresses specific challenges in infrastructure monitoring, demonstrating the comprehensive nature of our experimental framework.

In Figure 3, illustrative samples from the training set used in the experimental datasets are displayed. Table 3 provides a detailed overview of the four datasets employed, includ-ing the number of images in the training, test, and validation sets, the number of object classes and instances, and key descriptions, environments, and challenges that highlight the complexities facing the object detection models in real-world UAV-based applications.

Table 2. Challenge coverage across selected datasets.

| Challenge Category | InsPLAD-Det | Electric Substation | Common-VALID | Bird's Nest |
|---|---|---|---|---|
| Multi-scale Objects | ✓ | ✓ | ✓ | ✓ |
| Class Imbalance | ✓ | ✓ (Extreme) | ✓ | Single-class |
| Environmental Variability | ✓ | ✓ | ✓ | ✓ |
| Small Object Detection | ✓ | ✓ | ✓ | ✓ |
| Cluttered Backgrounds | ✓ | ✓ | ✓ | ✓ |
| Perspective Distortion | ✓ | ✓ | ✓ | ✓ |
| Defect/Anomaly Detection | ✓ | Limited | Limited | ✓ |
| Synthetic Data Integration | — | — | ✓ | — |

✓ indicates presence of the challenge; — indicates absence or not applicable.

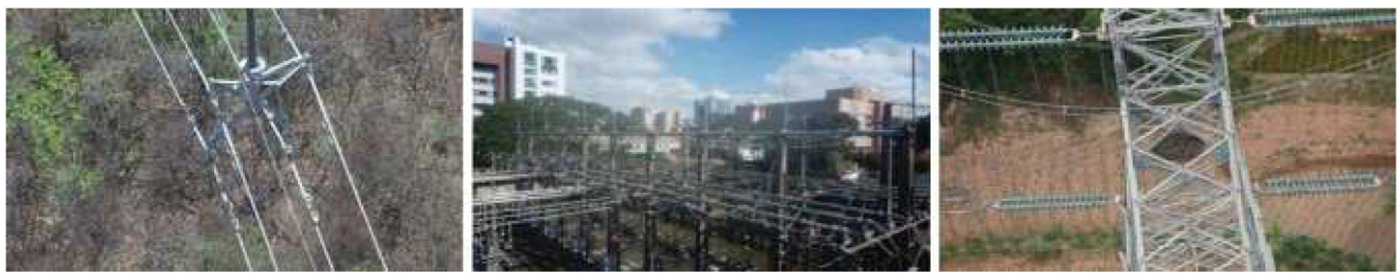

Figure 3. Representative samples from three evaluation datasets, highlighting the diversity of real-world object detection challenges. (**Left**) InsPLAD-det, showing power line infrastructure components with varying scales and complex backgrounds. (**Middle**) Electric Substation, featuring densely arranged objects and challenging lighting conditions. (**Right**) Bird's Nest dataset, capturing aerial views of nesting sites on electrical towers. These datasets collectively represent the target scenarios for deploying resource-efficient YOLO models in critical infrastructure monitoring.

Table 3. Summary of datasets used in research.

| Dataset | #Images Train | #Images Test | #Images Val | Classes | Instances | Description | Challenges |
|---|---|---|---|---|---|---|---|
| InsPLAD-det | 7907 | 2626 | 2626 [a] | 17 | 28,869 | UAV power line asset inspection in industrial environment [25] | Multi-scale objects, varied lighting, occlusions |
| Electric Substation | 5023 | 1452 | 601 | 15 | 202,509 | Electric substation images under diverse conditions in urban/industrial environment [26] | Varied weather/lighting, small objects |
| Common-VALID | 4246 | 908 | 900 | 8 | 70,799 | Complex environments for transfer learning in urban/remote environment [27] | Multi-class objects, small detection |
| Bird's Nest | 3000 [b] | 80 | 400 [b] | 1 | 3782 | Bird nest detection on electric towers in remote/wildlife environment [28] | High altitude, natural occlusions |

[a] The validation set is the same as the test set; [b] augmented data included in the final dataset splits.

## 5. Experimental Design

This section details the experimental methodology employed to systematically evaluate the impact of transfer learning strategies, particularly layer freezing, on the performance and computational efficiency of YOLOv8 and YOLOv10 models. Our general strategy involves two main components. We implement and compare several layer freezing configurations, where specific blocks of the pre-trained models' backbones are frozen during retraining on target datasets. These configurations, corresponding to freezing the initial 4 blocks (FR1), 9 blocks (FR2, encompassing the backbone), or 22/23 blocks (FR3) are benchmarked against standard fine-tuning and training from scratch. The evaluation across all experiments relies on key performance metrics (mAP@50, mAP@50:95) and resource indicators (GPU usage, training time), as described below.

*5.1. Layer Freezing and Implementation Details*

For our experiments, we will base the layer-freezing approach on the architectural details of the YOLO models under investigation (see Figures 1 and 2) and the percentage of parameters frozen. It is important to note that this number of parameters includes the cv2 and cv3 modules for YOLOv10, as the original paper introducing YOLOv10 does not include these parameters in Table 1 [13] since they are not needed during inference.

With regard to the parameters, the number of trainable parameters in the architecture's heads may vary, depending on the different number of object classes in the datasets discussed in Section 4. We will present the percentage of frozen parameters for each approach, model, and dataset, relative to the architecture of models pre-trained on the COCO [7] dataset, which includes 80 object classes, and including cv2 and cv3 modules for YOLOv10. However, the difference in the number of parameters between a model with 80 object classes and one with a single class is less than 0.3%. Table 4 shows the percentage of frozen parameters for each approach and model. "v8n-4b", "v8n-9b", and "v8n-22b" represent different approaches for the YOLOv8 model with variant "n" (nano), where "4b", "9b", and "22b" indicate the number of blocks frozen. Similarly, "v8s-4b", "v8s-9b", and "v8s-22b" refer to the "s" (small) variant of YOLOv8 with 4, 9, and 22 blocks frozen, respectively, and so forth for "m" (medium) and "l" (large) variants. The same structure is followed for the YOLOv10 model. These approaches involve freezing the layers enclosed in the areas labeled FR1, FR2, and FR3, as shown in Figures 1 and 2. Specifically, the FR1 regions correspond to freezing 4 blocks, the FR2 regions correspond to freezing 9 blocks, and the FR3 regions correspond to freezing 22 or 23 blocks for YOLOv8 and YOLOv10, respectively. For additional information on the selection of blocks to freeze, please refer to Appendix A.

Our selection of freezing configurations (FR1: 4 blocks; FR2: 9 blocks; FR3: 22/23 blocks) is grounded in the fundamental architectural boundaries of modern YOLO designs. Blocks 0–3 (FR1) constitute pure feature extraction layers positioned before any concatenation operations, capturing domain-agnostic visual representations (edges, textures, geometric patterns) with high transferability across detection tasks. The 9-block boundary (FR2) encompasses the complete backbone feature extraction pipeline, preserving all fundamental visual representations while allowing task-specific adaptation in the neck and head components. The extended 22/23-block configuration (FR3) includes most feature fusion operations, maintaining both feature extraction and integration capabilities while enabling fine-tuning only in the final detection layers. This hierarchical approach aligns with established transfer learning principles: early layers contain universal features suitable for preservation, intermediate layers handle feature integration requiring moderate adaptation, and final layers perform task-specific detection requiring full fine-tuning [5].

By selectively freezing the backbone blocks and layers in both YOLOv8 and YOLOv10 models, we aim to preserve the general feature extraction capabilities learned from the original dataset while allowing the upper layers to adapt to the specific characteristics of the new task. This approach will help us determine the optimal layers and blocks to freeze.

In our experiments, YOLOv8 and YOLOv10 models are trained using the stochastic gradient descent (SGD) optimizer for 1000 epochs, with an early stopping mechanism employing a patience of 30 epochs. The batch size is set to 16, and dataset images are cached in RAM to enhance training speed by minimizing disk I/O operations, which comes at the cost of increased memory usage. The momentum and weight decay for the SGD optimizer are configured to 0.937 and $5 \times 10^{-4}$, respectively. The initial learning rate is set to $1 \times 10^{-2}$, decaying linearly to $1 \times 10^{-4}$ over the course of training. All dataset images underwent standardized preprocessing to ensure consistency across experiments while preserving data distribution integrity. Images were resized to $640 \times 640$ pixels and normalized using ImageNet statistics (mean: [0.485, 0.456, 0.406]; standard deviation: [0.229, 0.224, 0.225]). We intentionally avoid applying data augmentation techniques during the training process to prevent introducing any stochastic shifts in the original distribution of the datasets, which could potentially affect the validity and reliability of our experimental results.

**Table 4.** Parameters frozen in each experimental approach.

| Model | Variant | #Param. (M) [a] | Approach | Frozen (%) |
|---|---|---|---|---|
| YOLOv8 | n | 3.2 | v8n-4b | 0.984 |
| | | | v8n-9b | 35.096 |
| | | | v8n-22b | 71.568 |
| | s | 11.2 | v8s-4b | 1.097 |
| | | | v8s-9b | 39.608 |
| | | | v8s-22b | 80.773 |
| | m | 25.9 | v8m-4b | 1.238 |
| | | | v8m-9b | 42.562 |
| | | | v8m-22b | 85.245 |
| | l | 43.7 | v8l-4b | 1.490 |
| | | | v8l-9b | 43.834 |
| | | | v8l-22b | 87.081 |
| YOLOv10 | n | 2.8 | v10n-4b | 1.119 |
| | | | v10n-9b | 28.268 |
| | | | v10n-23b | 66.500 |
| | s | 8.1 | v10s-4b | 1.507 |
| | | | v10s-9b | 27.570 |
| | | | v10s-23b | 79.076 |
| | m | 16.6 | v10m-4b | 1.935 |
| | | | v10m-9b | 38.488 |
| | | | v10m-23b | 85.681 |
| | l | 25.9 | v10l-4b | 2.515 |
| | | | v10l-9b | 51.374 |
| | | | v10l-23b | 88.623 |

[a] Note that this number of parameters includes the cv2 and cv3 modules for YOLOv10. In Table 1 [13], the number of parameters reported exclude these blocks, as they are not necessary during inference.

Additional hyperparameters include a warm-up phase consisting of three epochs, with a warm-up momentum of 0.8 and a warm-up bias learning rate of 0.1. The learning rate schedule follows a linear decay pattern. The specific loss gains are set as follows: the box loss gain is 7.5, the class loss gain is 0.5, and the dual focal loss gain is 1.5.

To ensure a fair comparison between architectures, all experiments for both YOLOv8 and YOLOv10 were conducted using the same hardware setup, an NVIDIA RTX A4000 GPU with 16 GB of memory. This unified configuration eliminates hardware-induced bias in training time and GPU memory usage metrics, enabling a direct and reliable comparison between the two architectures.

A comprehensive summary of the experimental environment and training hyperparameters is provided in Table 5.

Table 5. Experimental environment and hyperparameters.

| Parameter | Value |
| --- | --- |
| Training Configuration | |
| 　Optimizer | Stochastic gradient descents |
| 　Epochs | 1000 |
| 　Early stopping patience | 30 |
| 　Batch size | 16 |
| 　Image size | 640 × 640 |
| 　Initial learning rate (lr0) | $1 \times 10^{-2}$ |
| 　Final learning rate (lrf) | $1 \times 10^{-4}$ |
| 　Momentum | 0.937 |
| 　Weight decay | $5 \times 10^{-4}$ |
| 　Warmup epochs | 3 |
| 　Warmup momentum | 0.8 |
| 　Warmup bias LR | 0.1 |
| 　Box loss gain | 7.5 |
| 　Class loss gain | 0.5 |
| 　DFL loss gain | 1.5 |
| 　Data augmentation | Disabled |
| 　Seed | 42 |
| 　Number of runs | 1 per experiment |
| Validation Configuration | |
| 　Confidence threshold | 0.5 |
| 　NMS IoU threshold (YOLOv8) | 0.7 |
| Hardware Environment | |
| 　GPU | NVIDIA RTX A4000 (16 GB) |
| Software Environment | |
| 　Operating system | Ubuntu 22.04 LTS |
| 　Python | 3.10 |
| 　PyTorch | 2.0.1 |
| 　CUDA | 12.6 |

*5.2. Gradient Monitoring, Visual Analysis, and Metrics*

To gain deeper insights into the training dynamics and convergence behavior of different layer freezing strategies, we implement comprehensive gradient monitoring and visual analysis techniques. We aim to deepen the comprehension of one of the fundamental aspects of the proper convergence of the models during its training, the gradients. Concretely, the gradients of the model parameters with respect to the input. The formal definition of these parameters is as follows:

$$\nabla_\theta \mathcal{L}(\theta, x, y) = \left( \frac{\partial \mathcal{L}}{\partial \theta_1}, \frac{\partial \mathcal{L}}{\partial \theta_2}, \ldots, \frac{\partial \mathcal{L}}{\partial \theta_n} \right), \quad (1)$$

where $\theta = (\theta_1, \theta_2, \ldots, \theta_i)$ are the model parameters, $\mathcal{L}$ is the loss function, $x$ is the input, and $y$ is the target output.

For the monitoring of the gradients during training, we have decided to focus on their magnitudes using L2 Norm, defined as follows:

$$\|\nabla_\theta \mathcal{L}(\theta, x, y)\|_2 = \sqrt{\sum_{i=1}^{n} \left( \frac{\partial \mathcal{L}}{\partial \theta_i} \right)^2} \quad (2)$$

The L2 norm provides a scalar value representing the overall magnitude of the gradients for each layer or the entire model. By tracking the L2 norm over training iterations, we can gain insights into the training dynamics and identify potential issues. To visualize the L2 norm of gradients, we compute it at the end of each training batch.

The process begins by initializing a variable to accumulate the squared magnitudes of the gradients. For each parameter in the model, we check whether the gradient is available. If it is, we calculate the L2 norm of that gradient, square it, and add it to our accumulation variable. Once all parameters are processed, we take the square root of the accumulated value to obtain the L2 norm for that specific batch. This value is then stored in a list that tracks the L2 norms across all batches.

After the training process is completed for a specific epoch, we compute the mean of the L2 norms across all batches, excluding instances where the norm is zero to avoid skewing the average. This process provides a single representative value that reflects the overall gradient behavior during training. By using this approach, we can effectively monitor the gradients and assess the convergence of the model.

To complement the gradient magnitude analysis and provide visual understanding of model focus during training, we implement a gradient-weighted class activation mapping (GradCAM) analysis. This technique enables us to visualize which regions of the input images the model prioritizes during the detection process under different layer-freezing strategies.

The GradCAM methodology works by weighting the 2D activations by the average gradient, thereby generating visual explanations that highlight the key objects and regions that influence the model's predictions. For our YOLO architectures, we compute activation maps using the implementation provided by [32], which is specifically adapted for object detection models.

The GradCAM computation process involves several steps: first, we extract feature maps from a target convolutional layer within the model; second, we compute the gradients of the class score with respect to these feature maps; third, we pool the gradients across the width and height dimensions to obtain the neuron importance weights; and finally, we perform a weighted combination of the forward activation maps and apply a ReLU function to focus on features that have a positive influence on the class of interest.

This visual analysis methodology allows us to track how the model's attention evolves across training epochs and how different layer-freezing strategies affect the focus patterns. By generating activation maps at key training milestones (epoch 1, epoch 10, and the epoch with the best validation loss), we can observe the adaptation process and validate whether the models are learning to focus on relevant object features during the transfer learning process.

To thoroughly evaluate the performance of the YOLO models under different training strategies, we will focus on two key metrics: mAP@50 and mAP@50:95. These metrics provide a comprehensive assessment of the models accuracy and precision, with mAP@50 measuring the mean average precision at a 50% intersection over union (IoU) threshold and mAP@50:95 offering a more detailed evaluation across a range of IoU thresholds from 50% to 95%.

For the validation over the test and validation/test splits, we will use a minimum confidence threshold of 0.5. This setting ensures that only detections with a confidence level above 50% are considered, effectively discarding less certain detections and reducing the likelihood of false positives.

Additionally, we will apply an IoU threshold of 0.7 for non-maximum suppression (NMS) during the validation of YOLOv8 models. This threshold helps in reducing duplicate detections by suppressing overlapping bounding boxes, thereby improving the quality of the detected objects. It is important to note that IoU-based NMS is only applicable to YOLOv8, as YOLOv10 employs a free-NMS approach, which does not require a predefined IoU threshold for suppressing redundant detections.

By comparing the mAP@50 and mAP@50:95 values for different training strategies, including layer freezing, fine-tuning, and training from scratch, we can gain insights into the models' performance and the effectiveness of each approach in terms of accuracy and precision. Additionally, we will measure the training time and GPU usage for each case to assess the computational efficiency. The training time will be recorded in minutes, and GPU usage will be monitored using a tool like NVIDIA System Management Interface (nvidia-smi) to track the GPU memory consumption and utilization during the training process. The combination of accuracy metrics (mAP@50 and mAP@50:95) and computational efficiency metrics (training time and GPU usage) will provide a comprehensive evaluation of the YOLO models under different training strategies, enabling us to identify the most effective approach for balancing accuracy and computational resources.

## 6. Results

This section presents a comprehensive analysis of the experimental findings from various training strategies employed for YOLOv8 and YOLOv10 models. We delve into the primary outcomes in terms of performance metrics, examine the effects of layer freezing on gradient magnitudes and visual explanations using GradCAM, and provide additional insights into training time, resource utilization, and performance trade-offs. A detailed analysis of specific failure cases, particularly regarding the Bird's Nest dataset's anomalous behavior under aggressive layer freezing, is provided in Appendix C.

### 6.1. Performance Metrics and Trade-Offs

In this subsection, we detail the primary outcomes of our experiments, focusing on the performance metrics of the YOLOv8 and YOLOv10 models under different layer-freezing configurations. The tables below summarize the results, highlighting key metrics such as the mean average precision (mAP) at different IoU thresholds, the training time (in minutes), and the maximum GPU usage (in MB).

Table 6 presents the performance metrics for YOLOv8 across four different datasets (detailed in Section 4), illustrating how different freezing strategies impact the models' effectiveness and efficiency. Similarly, Table 7 provides the corresponding results for the YOLOv10 model on the same datasets. Each table includes various experimental approaches explained in Section 5.1, as well as the strategies of training from scratch and fine-tuning different model variants.

The performance metrics for YOLOv8 and YOLOv10, across different layer freezing strategies, reveal significant variations in training time and resource utilization. The training time is a critical factor when selecting an approach, as it impacts the practical feasibility of deploying and maintaining models, especially in resource-constrained environments.

Training time patterns vary considerably between YOLOv8 and YOLOv10. While YOLOv8 showed more consistent time reductions with fine-tuning compared to training from scratch, YOLOv10 exhibits mixed results. For YOLOv10, fine-tuning sometimes requires more time than training from scratch, particularly evident in the InsPLAD-det dataset where the large variant required 959 min for fine-tuning versus 404 min from scratch. However, on datasets like Electric Substation, fine-tuning generally reduces training time across model variants. The relationship between frozen layers and training time is also inconsistent in YOLOv10, with some configurations showing non-monotonic patterns where intermediate freezing strategies (e.g., v10n-9b) may take longer than more conservative approaches (e.g., v10n-4b).

Table 6. Performance metrics for YOLOv8 with different layer freezing strategies.

| Variant | Approach | InsPLAD-Det | | | | Electric Substation | | | | Common-VALID | | | | Bird's Nest | | | |
|---|---|---|---|---|---|---|---|---|---|---|---|---|---|---|---|---|---|
| | | GPU [a] | Time [b] | mAP$^{50}$ | mAP$^{50:95}$ | GPU [a] | Time [b] | mAP$^{50}$ | mAP$^{50:95}$ | GPU [a] | Time [b] | mAP$^{50}$ | mAP$^{50:95}$ | GPU [a] | Time [b] | mAP$^{50}$ | mAP$^{50:95}$ |
| n | From scratch | 2143 | 105 | 0.761 | 0.624 | 5175 | 60 | 0.752 | 0.486 | 3210 | 69 | 0.76 | 0.652 | 2193 | 98 | 0.807 | **0.403** |
| | Fine tuning | 2153 | 68 | 0.82 | 0.675 | 5135 | 39 | 0.786 | 0.518 | 3217 | 86 | 0.786 | 0.684 | 2216 | 87 | 0.83 | 0.372 |
| | v8n-4b | 1551 | 86 | 0.837 | 0.687 | 4555 | 35 | 0.778 | 0.512 | 2631 | 37 | 0.793 | 0.686 | 1637 | 16 | 0.78 | 0.353 |
| | v8n-9b | 1438 | 41 | 0.836 | 0.678 | 4286 | 39 | 0.762 | 0.49 | 2308 | 46 | 0.789 | 0.672 | 1287 | 72 | 0.788 | 0.378 |
| | v8n-22b | 901 | 53 | 0.594 | 0.446 | 3848 | 51 | 0.573 | 0.332 | 1897 | 67 | 0.726 | 0.578 | 801 | 40 | 0.4 | 0.184 |
| s | From scratch | 4238 | 118 | 0.761 | 0.619 | 6966 | 158 | 0.779 | 0.522 | 4785 | 262 | 0.766 | 0.677 | 4223 | 95 | 0.812 | 0.384 |
| | Fine tuning | 4294 | 129 | 0.842 | 0.712 | 6998 | 110 | 0.813 | 0.577 | 4871 | 98 | 0.811 | 0.724 | 4284 | 103 | 0.813 | 0.381 |
| | v8s-4b | 2961 | 125 | 0.84 | 0.707 | 5697 | 130 | 0.811 | 0.55 | 3556 | 61 | 0.816 | 0.725 | 2965 | 34 | 0.691 | 0.37 |
| | v8s-9b | 2281 | 153 | 0.835 | 0.695 | 5026 | 62 | 0.79 | 0.526 | 2728 | 100 | 0.809 | 0.713 | 2115 | 123 | 0.788 | 0.365 |
| | v8s-22b | 1237 | 143 | 0.684 | 0.516 | 4039 | 48 | 0.589 | 0.342 | 1864 | 101 | 0.757 | 0.62 | 1040 | 33 | 0.354 | 0.192 |
| m | From scratch | 6757 | 585 | 0.799 | 0.684 | 8973 | 285 | 0.798 | 0.552 | 6777 | 316 | 0.787 | 0.712 | 6729 | 220 | 0.802 | 0.359 |
| | Fine tuning | 6859 | 376 | 0.852 | 0.722 | 9072 | 313 | 0.82 | 0.574 | 6962 | 223 | 0.817 | 0.741 | 6851 | 270 | **0.839** | 0.385 |
| | v8m-4b | 4758 | 630 | 0.859 | **0.732** | 7241 | 278 | 0.819 | 0.57 | 4884 | 214 | 0.816 | 0.741 | 4630 | 64 | 0.694 | 0.377 |
| | v8m-9b | 3321 | 282 | 0.843 | 0.727 | 5765 | 139 | 0.811 | 0.557 | 3365 | 161 | 0.819 | 0.742 | 3338 | 152 | 0.761 | 0.382 |
| | v8m-22b | 1793 | 127 | 0.665 | 0.497 | 4223 | 121 | 0.613 | 0.359 | 1816 | 102 | 0.769 | 0.643 | 1713 | 133 | 0.547 | 0.231 |
| l | From scratch | 10,804 | 524 | 0.791 | 0.676 | 11,611 | 393 | 0.802 | 0.561 | 10,896 | 430 | 0.805 | 0.728 | 10,647 | 424 | 0.784 | 0.341 |
| | Fine tuning | 10,873 | 614 | **0.86** | 0.731 | 11,630 | 386 | 0.818 | 0.578 | 10,963 | 331 | 0.823 | 0.756 | 10,869 | 331 | 0.804 | 0.37 |
| | v8l-4b | 7262 | 268 | 0.854 | 0.725 | 8130 | 338 | **0.823** | **0.583** | 7163 | 193 | 0.829 | **0.757** | 7147 | 82 | 0.672 | 0.355 |
| | v8l-9b | 4724 | 408 | 0.85 | 0.73 | 5620 | 225 | 0.81 | 0.564 | 4785 | 170 | **0.831** | 0.756 | 4756 | 295 | 0.797 | 0.394 |
| | v8l-22b | 2384 | 245 | 0.727 | 0.538 | 3095 | 132 | 0.625 | 0.368 | 2537 | 175 | 0.778 | 0.663 | 2388 | 132 | 0.533 | 0.215 |

The best results per dataset are shown in bold; [a] the GPU usage is expressed as the maximum usage during training time in MB; [b] the time is expressed in minutes.

Table 7. Performance metrics for YOLOv10 with different layer freezing strategies.

| Variant | Approach | InsPLAD-Det | | | | Electric Substation | | | | Common-VALID | | | | Bird's Nest | | | |
|---|---|---|---|---|---|---|---|---|---|---|---|---|---|---|---|---|---|
| | | GPU [a] | Time [b] | mAP$^{50}$ | mAP$^{50:95}$ | GPU [a] | Time [b] | mAP$^{50}$ | mAP$^{50:95}$ | GPU [a] | Time [b] | mAP$^{50}$ | mAP$^{50:95}$ | GPU [a] | Time [b] | mAP$^{50}$ | mAP$^{50:95}$ |
| n | From scratch | 3207 | 177 | 0.658 | 0.513 | 6126 | 123 | 0.703 | 0.415 | 3806 | 87 | 0.608 | 0.497 | 3035 | 35 | 0.612 | 0.255 |
| | Fine tuning | 3175 | 222 | 0.824 | 0.646 | 6126 | 104 | 0.772 | 0.468 | 3808 | 85 | 0.690 | 0.591 | 3035 | 44 | 0.629 | 0.286 |
| | v10n-4b | 2527 | 151 | 0.824 | 0.648 | 5316 | 63 | 0.755 | 0.455 | 3441 | 99 | 0.693 | 0.587 | 2286 | 44 | 0.609 | 0.247 |
| | v10n-9b | 2080 | 247 | 0.801 | 0.627 | 5113 | 99 | 0.741 | 0.437 | 2965 | 46 | 0.686 | 0.564 | 1957 | 26 | 0.642 | 0.267 |
| | v10n-23b | 1418 | 140 | 0.45 | 0.317 | 4633 | 180 | 0.327 | 0.164 | 2420 | 127 | 0.470 | 0.359 | 1300 | 39 | 0.055 | 0.035 |
| s | From scratch | 5432 | 277 | 0.665 | 0.518 | 8567 | 225 | 0.742 | 0.452 | 5868 | 169 | 0.627 | 0.545 | 5247 | 51 | 0.669 | 0.268 |
| | Fine tuning | 5442 | 348 | 0.827 | 0.665 | 8449 | 116 | 0.809 | 0.508 | 5801 | 94 | 0.705 | 0.614 | 5247 | 69 | 0.633 | 0.264 |
| | v10s-4b | 4182 | 236 | 0.833 | 0.669 | 6933 | 70 | 0.798 | 0.497 | 4773 | 81 | 0.709 | 0.622 | 3995 | 53 | 0.741 | 0.343 |
| | v10s-9b | 3097 | 278 | 0.824 | 0.667 | 5836 | 76 | 0.783 | 0.482 | 3792 | 122 | 0.711 | 0.627 | 3047 | 36 | 0.72 | 0.307 |
| | v10s-23b | 1923 | 253 | 0.512 | 0.359 | 4706 | 70 | 0.392 | 0.196 | 2668 | 98 | 0.538 | 0.410 | 1883 | 51 | 0.192 | 0.071 |
| m | From scratch | 9337 | 564 | 0.691 | 0.563 | 11587 | 269 | 0.75 | 0.464 | 9469 | 309 | 0.654 | 0.581 | 9123 | 86 | 0.632 | 0.275 |
| | Fine tuning | 9337 | 601 | 0.829 | 0.669 | 11490 | 155 | 0.822 | 0.525 | 9471 | 215 | 0.717 | 0.645 | 9123 | 90 | 0.641 | 0.279 |
| | v10m-4b | 6868 | 282 | 0.828 | 0.674 | 8974 | 120 | 0.815 | 0.516 | 6939 | 151 | 0.718 | 0.639 | 6459 | 54 | 0.727 | 0.308 |
| | v10m-9b | 4796 | 125 | 0.824 | 0.652 | 7105 | 101 | 0.812 | 0.509 | 5029 | 100 | 0.733 | 0.645 | 4656 | 34 | 0.753 | 0.321 |
| | v10m-23b | 2611 | 285 | 0.507 | 0.362 | 5140 | 223 | 0.463 | 0.236 | 2835 | 320 | 0.578 | 0.461 | 2489 | 64 | 0.227 | 0.093 |
| l | From scratch | 13,518 | 404 | 0.654 | 0.537 | 13,847 | 341 | 0.741 | 0.46 | 13,619 | 300 | 0.647 | 0.573 | 13,185 | 136 | 0.624 | 0.263 |
| | Fine tuning | 13,491 | 959 | 0.825 | 0.664 | 13,839 | 152 | 0.83 | 0.528 | 13,590 | 298 | 0.716 | 0.642 | 13,183 | 90 | 0.676 | 0.277 |
| | v10l-4b | 10,576 | 194 | 0.83 | 0.653 | 10,891 | 126 | **0.832** | **0.529** | 10,473 | 349 | 0.735 | **0.669** | 10,255 | 60 | **0.779** | 0.303 |
| | v10l-9b | 7225 | 482 | **0.833** | **0.671** | 7793 | 107 | 0.816 | 0.514 | 7485 | 311 | **0.739** | 0.666 | 7357 | 54 | 0.748 | **0.349** |
| | v10l-23b | 3003 | 385 | 0.543 | 0.381 | 3901 | 148 | 0.483 | 0.249 | 3284 | 320 | 0.602 | 0.485 | 2928 | 109 | 0.29 | 0.113 |

The best results per dataset are shown in bold; [a] the GPU usage is expressed as the maximum usage during the training time in MB; [b] the time is expressed in minutes.

GPU usage, another crucial metric, decreased with an increase in the number of frozen layers. This reduction is due to fewer parameters being updated during training, which lowers computational demands. For example, the "v8n-22b" approach in YOLOv8 showed a significant reduction in GPU usage compared to both training from scratch and fine-tuning, with values dropping from 2143 MB to 901 MB for the InsPLAD-det dataset. A similar trend was noted for YOLOv10, where "v10l-23b" used markedly less GPU memory than the baseline models.

The trade-offs between resource efficiency and model performance are evident in the results. While freezing more layers leads to lower GPU usage and usually shorter training times, it also typically results in lower mAP scores. For example, the YOLOv8 "v8n-22b" and YOLOv10 "v10n-23b" variants, despite their efficient resource utilization, suffered from significantly lower mAP scores across all datasets compared to other approaches like fine-tuning and backbone-only freezing. Notably, the Bird's Nest dataset exhibited catastrophic failure with the "v10n-23b" configuration, achieving only 0.055 mAP@50 compared to 0.629 with fine-tuning. A detailed analysis of this failure mode is presented in Appendix C. In contrast, approaches that only freeze the backbone layers (e.g., "v8n-9b" and "v10s-9b") strike a balance by offering a moderate reduction in resource usage without a substantial drop in mAP scores. These variants achieved respectable mAP scores while still benefiting from reduced GPU usage and training times, making them a practical choice for applications requiring a balance between performance and efficiency. In contrast, approaches that only freeze the backbone layers (e.g., "v8n-9b" and "v10s-9b") strike a balance by offering a moderate reduction in resource usage without a substantial drop in mAP scores.

To synthesize these findings, Table 8 presents the optimal layer freezing configurations for each dataset, highlighting the best-performing approaches that balance accuracy with computational efficiency. To produce these synthesized findings, the results from the YOLOv8 and YOLOv10 experiments are combined into a single pool, ensuring comparability since all experiments are executed on the same hardware setup. Then, for each dataset, configurations are grouped and analyzed using mAP@50 as the primary optimization criterion. The selection process follows a two-stage filtering approach: first, the absolute best-performing configuration per dataset is identified based on mAP@50. Subsequently, a performance tolerance threshold of 1.5% is applied to create a pool of acceptable configurations that maintain performance within this margin of the optimal value. Among configurations meeting the performance criteria, the final recommendation prioritizes computational efficiency by selecting the configuration with minimal GPU memory usage. This approach balances performance preservation with resource optimization, acknowledging that marginal performance differences often do not justify significantly higher computational costs. Performance decreases and GPU savings are calculated relative to the best-performing configuration to quantify the efficiency gains achieved through this methodology.

**Table 8.** Optimal layer-freezing configurations by dataset.

| Dataset | Recommendation | mAP@50 | mAP@50:95 | GPU [a] | GPU Savings | Performance Drop |
|---|---|---|---|---|---|---|
| Bird's Nest | YOLOv8-n Fine tuning | 0.830 | 0.372 | 2216 | 68% | 1.1% |
| Common-VALID | YOLOv8-m FR2 | 0.819 | 0.742 | 3365 | 30% | 1.4% |
| Electric Substation | YOLOv8-l FR1 | 0.823 | 0.583 | 8130 | 25% | 1.1% |
| InsPLAD-det | YOLOv8-l FR2 | 0.850 | 0.730 | 4724 | 57% | 1.2% |

Note: GPU savings and performance drop are calculated relative to the highest-performing configuration for each dataset (within 1.5% tolerance); [a] GPU usage is expressed as the maximum usage during the training time in MB.

Our enhanced analysis, considering all training approaches (from scratch, fine-tuning, FR1, FR2, and FR3) and optimizing for mAP@50, reveals that strategic method selection achieves optimal efficiency across datasets. The recommended configurations show fine-tuning in 25% of cases, FR1 in 25% of cases, and FR2 in 50% of cases.

When resource-efficient alternatives are selected over the highest-performing configurations, they achieve an average GPU memory reduction of 45% with a minimal performance impact (average drop of 1.2% in mAP@50). This analysis demonstrates that carefully selected layer freezing strategies and model variants can significantly reduce computational requirements while maintaining competitive performance across diverse object detection tasks when optimizing for mAP@50. Additionally, the correlation between mAP@50 and mAP@50:95 across all configurations is 0.740, indicating moderate alignment between the two metrics. This suggests that optimizing for mAP@50 generally preserves performance in mAP@50:95.

For the sake of visualizing this balance, Figure 4 and Figure 5 illustrate the trade-off between maximum GPU usage and mAP50 obtained in the different experimental approaches and datasets for YOLOv8 and YOLOv10, respectively. It can be observed that the best balance, for most of the datasets, is achieved when freezing the first four blocks or the full backbone of the models.

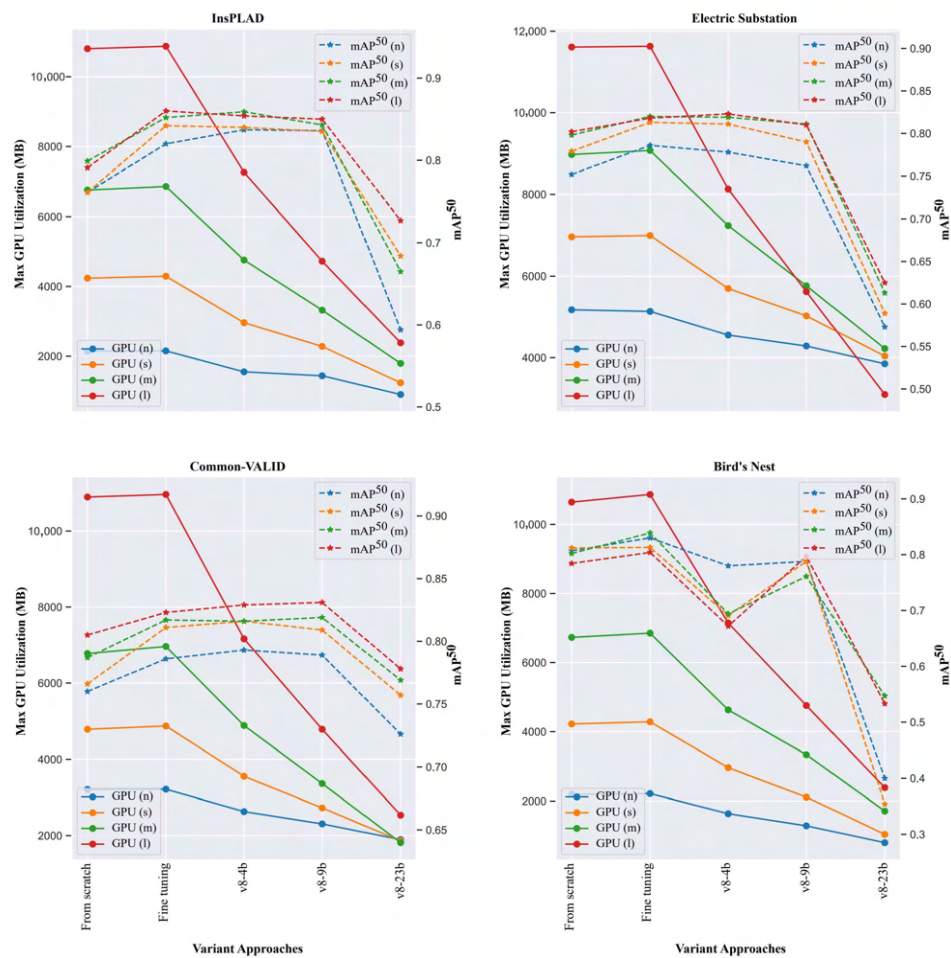

**Figure 4.** Trade-off between GPU memory usage and mAP@50 for various YOLOv8 layer freezing strategies. Each point denotes a specific configuration (model, dataset, and approach). Moderate freezing, v8-4b (first 4 blocks) and v8-9b (backbone), offers the best balance of accuracy and efficiency. Aggressive freezing (v8-22b) minimizes resources but reduces accuracy, while full fine-tuning and training from scratch require the most resources with mixed accuracy gains.

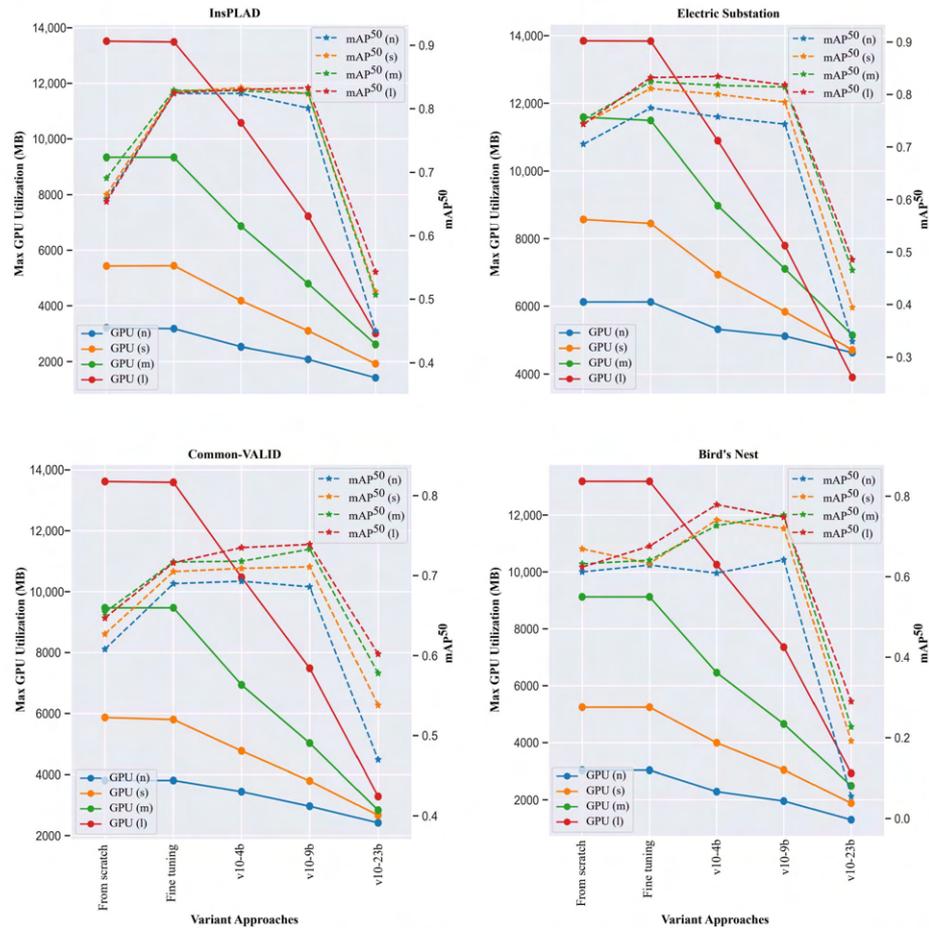

**Figure 5.** Performance-efficiency trade-off for YOLOv10 architectures, highlighting the relationship between GPU usage and mAP@50. Similar to YOLOv8, optimal trade-offs occur with moderate freezing (v10-4b, v10-9b), reducing memory use by up to 28% with minimal accuracy loss.

### 6.2. Gradient Behavior and Visual Analysis

Using the gradient monitoring methodology described in Section 5.2, we analyze the training dynamics across different layer freezing strategies.

Figure 6 shows the evolution of the L2 norm gradients for the four datasets of the small-sized models of YOLOv10.

During initial training phases, the L2 gradient norm is elevated but progressively diminishes in three out of four datasets, reflecting significant early updates that slow as the model converges. This decrease is irregular across datasets, highlighting the complex and rugged nature of the loss landscape, with the model encountering saddle points and local minima. In fine-tuning, the L2 norm is initially lower compared to training from scratch, as fine-tuning starts with a pre-trained model requiring smaller updates. The Electric Substation dataset shows more divergent or increasing gradients over epochs, likely due to the highly variable atmospheric conditions and the strong bias towards the "Porcelain Pin Insulator" class. Freezing various parts of the model results in distinct gradient behaviors: freezing four blocks (v10s-4b) results in moderate L2 norms, indicating substantial network adaptation; freezing the backbone (v10s-9b) shows further reduced L2 norms, suggesting fewer changes in early layers; and the last freezing approach (v10s-23b) leads to the second-highest L2 norms, specifically in the Electric Substation, Common-VALID, and Bird's Nest datasets, with most weights unchanged and only the final layers adapting.

To provide a comprehensive quantitative validation of the observed gradient behaviors, Table 9 presents statistical analysis of gradient magnitudes across all training strategies and datasets. The analysis reveals distinct optimization regimes: training from scratch exhibits the highest gradient magnitudes with extreme variability (82–99% coefficient of variation), indicating chaotic optimization landscapes. Fine-tuning and partial freezing strategies demonstrate significantly improved stability, with backbone freezing (v10s-9b) achieving optimal balance through 10–15% magnitude reduction compared to fine-tuning while maintaining low variability. The most constrained approach (v10s-23b) shows dramatically reduced gradients (16K–96K range), confirming concentrated learning in detection heads.

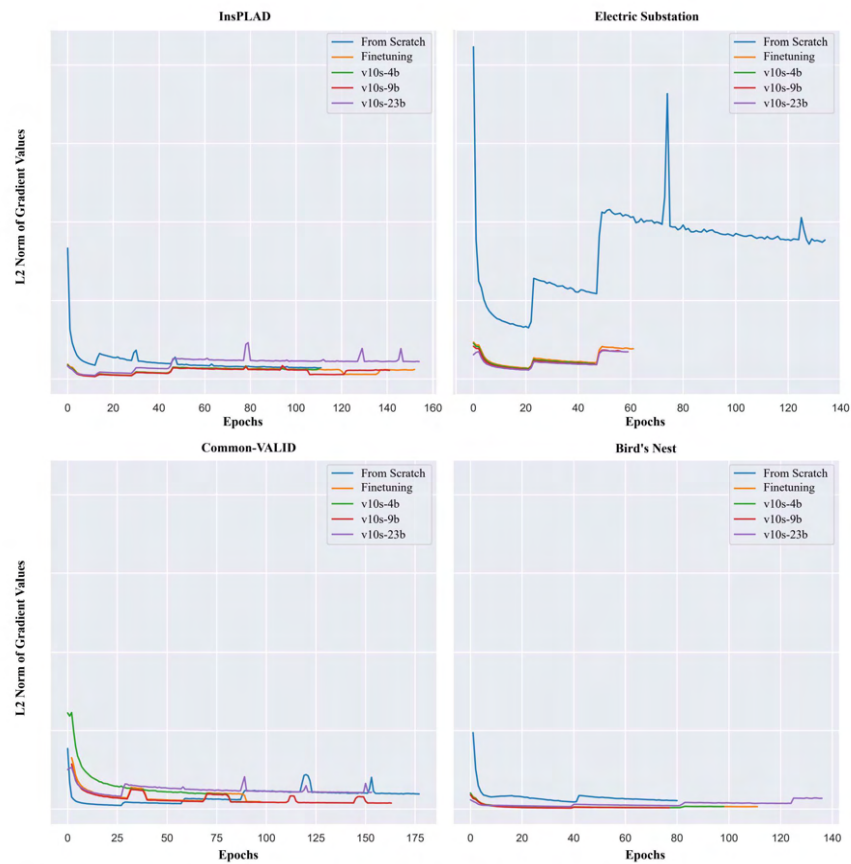

**Figure 6.** Evolution of L2 norm gradients during training for YOLOv10-small across four datasets, revealing distinct convergence patterns under different transfer learning strategies. Training from scratch exhibits high initial gradients that gradually decrease, while fine-tuning starts with lower gradients due to pre-trained initialization. Layer freezing approaches show varying gradient magnitudes: v10s-4b maintains moderate adaptation, v10s-9b demonstrates reduced backbone updates, and v10s-23b concentrates learning in final layers. The Electric Substation dataset's divergent behavior reflects dataset complexity and class imbalance challenges, highlighting the importance of gradient monitoring for transfer learning optimization.

It can also be observed that, after several epochs, the L2 norm of the gradients does not consistently decrease and, in some instances, even increases, despite the loss function continuing to decrease. The fluctuations in the L2 norm of gradients during training can be attributed to several factors; some of them may be as follows.

- Learning rate dynamics: The initial learning rate of $1 \times 10^{-2}$, while effective for rapid convergence, may also lead to instability in gradient updates, especially in the presence of momentum. The instability arises because the momentum term accumulates previous gradients, which, combined with a high learning rate, can lead to overshooting the minima and cause divergent behavior. In the context of optimization algorithms, a high learning rate can cause the gradients to oscillate, rather than converge smoothly. This issue is particularly pronounced when momentum is used, as it can amplify the oscillations. The instability arises because the momentum term accumulates previous gradients, which, combined with a high learning rate, can lead to overshooting the minima and cause divergent behavior. This phenomenon is described in [33,34], where the authors emphasize that the relationship between learning rate and momentum must be carefully managed to prevent instability and oscillations, which are more likely at higher learning rates.

- Loss landscape and saddle points: The complexity of navigating saddle points in non-convex optimization landscapes is well documented in the literature [35–37]. These landscapes can lead to temporary increases in gradient magnitudes even as the overall loss decreases. This complicates the interpretation of convergence and makes optimization more challenging. Gradient descent methods can experience slow convergence or even stall at saddle points due to the nature of the Hessian at these points. The eigenvalues of the Hessian near saddle points can be both positive and negative, causing the gradients to point in directions that do not uniformly decrease the loss across all dimensions.

Table 9. Gradient magnitude statistics across training strategies and datasets, showing mean gradient norms, standard deviations, and coefficients of variation. The analysis demonstrates distinct optimization regimes from chaotic (scratch training) to highly magnitude-constrained (v10s-23b).

| Dataset | Strategy | Mean Gradient Norm | Standard Deviation | Coefficient of Variation (%) |
|---|---|---|---|---|
| Birds-Nest | From Scratch | 140,671 | 122,384 | 87.0 |
|  | Fine tuning | 27,174 | 9339 | 34.4 |
|  | v10s-4b | 25,567 | 8924 | 34.9 |
|  | v10s-9b | 23,948 | 7481 | 31.2 |
|  | v10s-23b | 16,227 | 5443 | 33.5 |
| Common-VALID | From Scratch | 119,044 | 117,309 | 98.5 |
|  | Fine tuning | 231,463 | 37,172 | 16.1 |
|  | v10s-4b | 218,001 | 37,765 | 17.3 |
|  | v10s-9b | 203,452 | 32,473 | 16.0 |
|  | v10s-23b | 96,119 | 15,612 | 16.2 |
| Electric-Substation | From Scratch | 216,028 | 191,936 | 88.9 |
|  | Fine tuning | 79,508 | 14,667 | 18.5 |
|  | v10s-4b | 75,193 | 15,085 | 20.1 |
|  | v10s-9b | 70,012 | 14,165 | 20.2 |
|  | v10s-23b | 58,991 | 10,226 | 17.3 |
| InsPLAD | From Scratch | 140,271 | 115,582 | 82.4 |
|  | Fine tuning | 27,839 | 7806 | 28.0 |
|  | v10s-4b | 25,793 | 8032 | 31.1 |
|  | v10s-9b | 23,674 | 8357 | 35.3 |
|  | v10s-23b | 23,527 | 7321 | 31.1 |

To provide visual validation of our findings, Figure 7 displays activation maps generated from a sample image in the Common-VALID dataset using the GradCAM methodology detailed in Section 5.2. These maps, computed at epochs 1 and 10, as well as the best validation epoch, demonstrate how different training strategies affect the model's attention patterns.

The fine-tuning and freezing approach began with pre-trained weights from the COCO dataset, which includes the "car" class among its 80 classes. This prior knowledge enabled the model to detect vehicles with relatively high confidence, exceeding 80%, after a single epoch. As training progressed, the model increasingly focused its activations on the vehicles, indicating that the model parameters were adapting appropriately to the new dataset. The activation maps were computed using [32], which helps in visualizing the key objects that a model, such as YOLOv10, focuses on within an image. This method works by weighting the 2D activations by the average gradient, thereby offering insights into the model behavior during the training process.

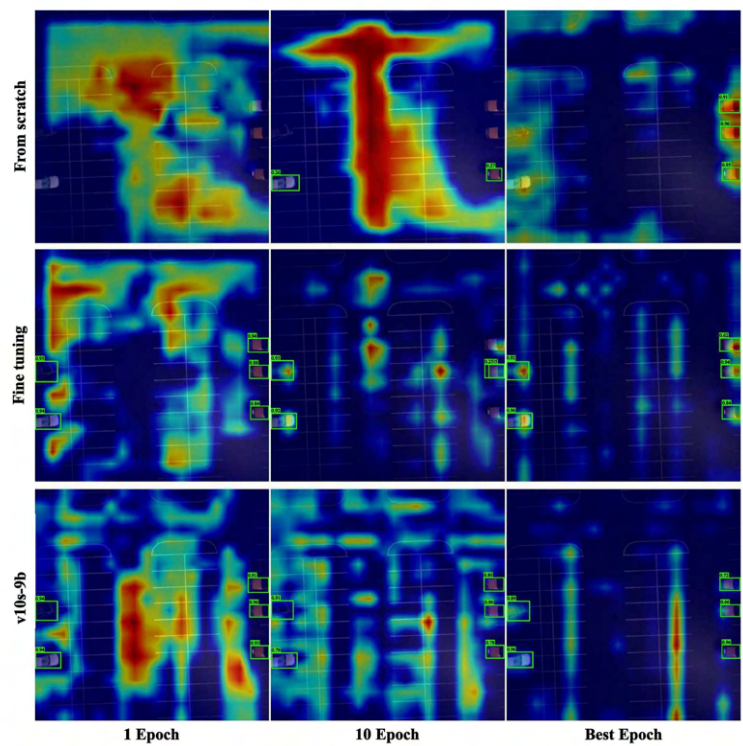

**Figure 7.** Temporal evolution of gradient-weighted class activation maps (GradCAM) for YOLOv10-small, demonstrating attention pattern development across training epochs. Three training strategies are compared: training from scratch, fine-tuning, and backbone freezing (v10s-9b). Maps are generated at epochs 1 and 10, as well as the best validation performance, showing a progressive refinement in the object focus. The color intensity represents activation magnitude, where red indicates higher activation norms and blue indicates lower activation norms. The pre-trained approaches (fine-tuning and v10s-9b) leverage COCO dataset knowledge to achieve immediate vehicle detection confidence higher than 80%, while training from scratch requires more epochs to develop comparable attention patterns. This visualization validates the effectiveness of transfer learning in preserving and adapting learned feature representations.

## 7. Discussion

Our comprehensive analysis reveals the fundamental mechanisms underlying the effectiveness of layer freezing strategies in YOLO architectures. The superior performance of FR1 (4-block) and FR2 (backbone) configurations stems from the hierarchical nature of feature learning in convolutional neural networks, where early layers capture universal visual features like edges and textures, while later layers develop task-specific representations. Gradient analysis demonstrates that these strategies create distinct optimization dynamics, with FR2 emerging as the most robust approach across diverse scenarios by balancing feature preservation and adaptation flexibility. In contrast, FR3 offers extreme efficiency but at the cost of reduced adaptation capacity.

Comparing architectures, YOLOv8 generally achieves higher mAP@50 and mAP@50:95 scores across most datasets and configurations, demonstrating superior baseline performance. However, both YOLOv8 and YOLOv10 benefit similarly from strategic layer freezing, with FR1 and FR2 consistently outperforming more aggressive strategies. This suggests that the principles of hierarchical feature preservation apply effectively across different YOLO architectures, regardless of specific innovations.

Results reveal distinct interactions between freezing strategies and the unique characteristics of each dataset, directly informing the optimal choices for practical infrastructure monitoring use cases, as summarized in Table 8. For the Bird's Nest dataset, a single-class detection task is characterized by heavy data augmentation and fine-grained texture discrimination, full fine-tuning proved optimal. The significant domain shift introduced by augmentation required full model adaptability, which more restrictive freezing strategies could not provide. Conversely, the Electric Substation dataset, defined by extreme class imbalance and dense, small objects, benefited most from the FR1 (4-block freeze) strategy. This approach allows for adaptation in the mid-to-late layers, which is critical for learning the complex distribution of varied equipment types under diverse environmental conditions, while preserving only the most foundational features. For the InsPLAD-det and Common-VALID datasets, which feature multiple classes of common objects in contextually rich scenes, the FR2 (backbone freeze) strategy emerged as superior. Freezing the entire backbone leverages the robust, general-purpose features learned from the COCO dataset, while allowing the neck and head to specialize in detecting specific assets and vehicle types. These findings demonstrate that the choice of an optimal freezing strategy is not universal but is instead a direct function of the target application's data properties, such as class diversity, object density, and the presence of augmentation.

Beyond accuracy, strategic freezing yields significant efficiency gains, with FR1 and FR2 reducing GPU memory consumption by up to 28% and 44%, respectively, and significant training time savings compared to training from scratch. The non-linear savings in training times highlight how architectural design influences efficiency, enabling rapid model iteration for UAV-based infrastructure monitoring under power and processing constraints. The competitive performance of FR1 and FR2 challenges conventional transfer learning approaches, suggesting that selective freezing enhances both accuracy and efficiency by preventing overfitting while preserving generalizable features.

These findings extend prior work on layer freezing in earlier YOLO variants, such as YOLOv5, by demonstrating size-dependent optimization strategies across modern architectures and diverse real-world datasets. While previous studies focused primarily on backbone freezing for specific tasks like vehicle detection, our analysis reveals nuanced guidelines: smaller models (nano and small variants) require more adaptive approaches like FR1 to maintain performance, whereas larger models tolerate aggressive freezing due to their greater parameter capacity. This size-dependent behavior underscores the importance of model scale in transfer learning design, providing a framework for practitioners to select appropriate strategies based on available computational resources and deployment requirements.

Furthermore, the integration of gradient monitoring with visual explanations offers a diagnostic methodology that bridges quantitative performance metrics with interpretable insights into model behavior. This approach not only validates the effectiveness of moderate freezing strategies but also highlights potential failure modes, such as the erratic gradient patterns in imbalanced datasets under aggressive freezing. By correlating these dynamics with final performance outcomes, our work contributes to a deeper theoretical understanding of transfer learning in object detection, potentially informing the development of automated freezing optimization tools that adapt to specific dataset characteristics and architectural constraints.

Several limitations must be acknowledged. Our static freezing approach may be suboptimal compared to dynamic methods. The absence of edge device validation limits applicability to real-world UAV deployments, where environmental consistency is crucial.

Future research should explore adaptive freezing algorithms, hybrid optimizations like quantization and pruning, and actual deployments on platforms like NVIDIA Jetson to validate inference performance and power benefits.

## 8. Conclusions

This study demonstrates that layer freezing in YOLO architectures effectively optimizes the trade-off between computational efficiency and detection performance, addressing key gaps in prior research that primarily focused on older architectures like YOLOv5 with limited dataset diversity and insufficient analysis of training dynamics. Through a systematic evaluation across modern YOLOv8 and YOLOv10 variants on four diverse real-world datasets representing critical infrastructure monitoring, our findings extend beyond basic efficiency–accuracy balancing by providing insights into size-dependent freezing strategies and dataset-specific adaptation patterns. Key contributions include actionable deployment guidelines, such as backbone freezing for an optimal performance–efficiency balance in most scenarios, four-block freezing for severely constrained resources, and gradient monitoring for the early detection of training instabilities, along with a gradient-based analysis methodology that equips practitioners with diagnostic tools for optimizing freezing strategies in various contexts.

These advancements enable advanced object detection on edge devices and accelerate model iteration for UAV-based infrastructure monitoring, addressing key barriers in resource-constrained environments.

Limitations include our static freezing approach and lack of edge device validation. Future research should develop adaptive freezing algorithms that dynamically adjust frozen parameters based on training dynamics, such as gradient flow, layer-wise feature relevance, or domain shift indicators, potentially using real-time feedback from validation metrics or reinforcement learning. Comprehensive deployment studies on actual UAV platforms are essential to assess practical benefits, including latency, energy consumption, and environmental robustness.

**Author Contributions:** Conceptualization, A.D.D. and A.M.B.; methodology, J.R.C.; software, A.D.D.; validation, A.D.D. and A.M.B.; formal analysis, A.D.D.; investigation, A.M.B.; resources, J.R.C.; data curation, A.D.D.; writing—original draft preparation, A.D.D. and A.M.B.; writing—review and editing, J.R.C.; visualization, A.D.D.; supervision, J.R.C.; project administration, A.M.B.; funding acquisition, J.R.C. All authors have read and agreed to the published version of the manuscript.

**Funding:** The authors acknowledge funding under grant 01103386 by the Horizon Europe EDF Programme, and under grant MIA.2021.M04.0008 by the Spanish Ministry of Economic Affairs and Digital Transformation and under national grant PID2023-151605OB-C21 by the Spanish Ministry of Science and Innovation.

**Institutional Review Board Statement:** Not applicable.

**Informed Consent Statement:** Not applicable.

**Data Availability Statement:** The data that support the findings of this study are openly available in the following repositories. InsPLAD-det: https://github.com/andreluizbvs/InsPLAD/tree/main, accessed on 27 July 2025. Electric Substation: https://figshare.com/articles/dataset/A_YOLO_Annotated_15-class_Ground_Truth_Dataset_for_Substation_Equipment/24060960, accessed on 27 July 2025. VALID: https://sites.google.com/view/valid-dataset, accessed on 27 July 2025. Birds Nest: https://zenodo.org/records/4015912#.X1O_0osRVPY, accessed on 27 July 2025. Additionally, the code necessary to replicate the metrics, as well as the exact datasets used in this research, are available at the HuggingFace repository: https://huggingface.co/AndrzejDD/enhanced_transfer_learning, accessed on 27 July 2025.

**Conflicts of Interest:** The authors declare no conflicts of interest.

# Appendix A. Implications of SPPF and PSA on Layer Freezing

During our experimental approaches' definition, we encountered the question of whether, when freezing the backbone (nine blocks for both models), we should include the SPPF block and the PSA block in the case of YOLOv10. It is well known that the backbone of these models acts as a feature extractor, capturing low-level characteristics in the initial stages (such as edges, corners, and small patches of a given image), and further layers combine these details to detect objects per se.

The SPPF layer pools the features and generates fixed-length outputs, performing information aggregation at a deeper stage of the network hierarchy. Due to this role, its implication and the necessity of fitting its trainable parameters during training concerning the backbone remained unclear. The PSA block, which enhances performance by integrating global modeling capabilities with a minimal computational cost, also posed a similar question. The PSA block is known for its efficiency in handling attention mechanisms by reducing redundancy in attention heads, thus providing a balanced trade-off between accuracy and computational overhead.

To address these uncertainties, we conducted a constrained experiment in which we fit the small (s)-size models on the common-VALID and Electric Substation datasets, freezing either the backbone or the backbone + SPPF in the case of YOLOv8, and the backbone or backbone + SPPF + PSA for YOLOv10. The results in Tables A1 and A2 highlight the performance metrics of YOLOv8 and YOLOv10 models using different layer freezing strategies across the Common-VALID and Electric Substation datasets. Notably, the strategies that freeze only the backbone layers (v8s-9b and v10s-9b) offer a balanced approach in terms of performance and resource utilization. In the Common-VALID dataset, YOLOv8 (v8s-9b) achieves mAP@50 of 0.881 and mAP@50:95 of 0.791, while YOLOv10 (v10s-9b) reaches mAP@50 of 0.713 and mAP@50:95 of 0.618. For the Electric Substation dataset, YOLOv8 (v8s-9b) obtains mAP@50 of 0.791 and mAP@50:95 of 0.526, whereas YOLOv10 (v10s-9b) achieves mAP@50 of 0.784 and mAP@50:95 of 0.483.

Although freezing more layers (such as the SPPF layer for YOLOv8 and SPPF+PSA for YOLOv10) results in lower GPU usage, it leads to decreased mAP scores. The backbone-only freezing (v8s-9b and v10s-9b) achieves a better trade-off between model performance and computational efficiency, retaining higher mAP scores while still being resource-efficient. This strategy maintains essential learned features while allowing enough flexibility for the model to adapt to new data. The decision to exclude the SPPF and PSA blocks from freezing is based on their roles in feature pooling and enhancing global representation learning. By not freezing these blocks, we maintain optimal feature extraction and attention capabilities, with only a slight additional computational overhead. This proves to be an effective approach for these datasets.

Table A1. Performance metrics for Common-VALID Dataset with additional layer freezing strategies.

| Model | Variant | Approach | Frozen (%) | GPU [a] | Time [b] | mAP$^{50}$ | mAP$^{50:95}$ |
|---|---|---|---|---|---|---|---|
| YOLOv8 | s | v8s-9b | 39.608 | 2743 | 94 | 0.881 | 0.791 |
|  |  | v8s-10b | 45.490 | 2634 | 63 | 0.794 | 0.698 |
| YOLOv10 |  | v10s-9b | 27.570 | 3848 | 50 | 0.713 | 0.618 |
|  |  | v10s-11b | 47.843 | 3525 | 81 | 0.695 | 0.603 |

[a] GPU usage is expressed as the maximum usage during training time in MB; [b] time is expressed in minutes.

**Table A2.** Performance metrics for Electric Substation Dataset with additional layer-freezing strategies.

| Model | Variant | Approach | Frozen (%) | GPU [a] | Time [b] | mAP$^{50}$ | mAP$^{50:95}$ |
|---|---|---|---|---|---|---|---|
| YOLOv8 | s | v8s-9b | 39.608 | 5026 | 62 | 0.79 | 0.526 |
|  |  | v8s-10b | 45.490 | 4655 | 38 | 0.78 | 0.518 |
| YOLOv10 |  | v10s-9b | 27.570 | 5943 | 33 | 0.784 | 0.483 |
|  |  | v10s-11b | 47.843 | 5712 | 31 | 0.761 | 0.459 |

[a] GPU usage is expressed as the maximum usage during training time in MB; [b] time is expressed in minutes.

## Appendix B. Dataset Distributions

*Appendix B.1. Electric Substation*

In this appendix, we provide additional information about the Electric Substation dataset, including the distribution of images and instances across the 15 object classes. Two histograms are included to illustrate this distribution.

The Electric Substation dataset consists of 7539 images, which are labeled with 213,566 annotations across 15 different object classes. The dataset was carefully curated to remove a small percentage of unlabeled images, resulting in a dataset with a rich diversity of labeled instances. The total number of unique images in the dataset is 7076.

The histograms below illustrate the distribution of instances and images across the 15 object classes in the Electric Substation dataset. Figure A1 shows the number of images containing each class, with most classes appearing in nearly 2000 images. The total count of images across all classes is 34,728, which is higher than the total number of unique images (7076). This discrepancy indicates that various classes often appear in the same image, resulting in multiple annotations per image.

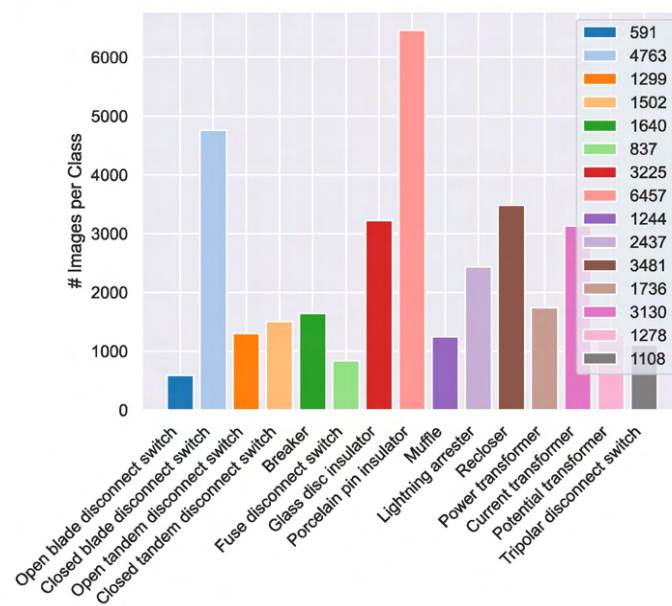

**Figure A1.** Histogram showing the number of images per class in the Electric Substation dataset.

Figure A2 displays the number of instances for each class in the dataset, highlighting a strong bias towards the "Porcelain pin insulator" class, which has 107,497 instances across the 7076 images.

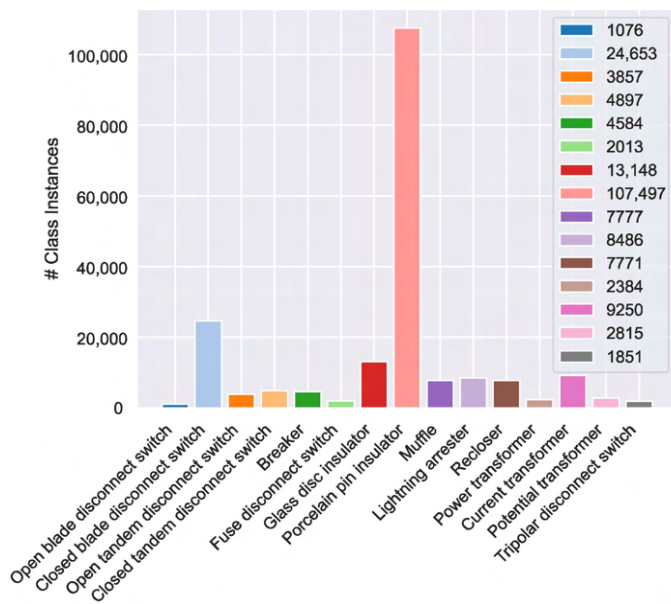

**Figure A2.** Histogram showing the number of instances per class in the Electric Substation dataset.

*Appendix B.2. Common-VALID*

In this appendix, we provide detailed information about the Common-VALID dataset, which was curated by selecting the most representative classes for common transfer learning scenarios based on the VALID dataset. This streamlined dataset facilitated the evaluation of the effects of freezing layers in various YOLO architectures, comparing them to training from scratch or fine-tuning the models.

The histograms below illustrate the distribution of instances and images across the 8 object classes in the Common-VALID dataset. These histograms provide a visual representation of the data distribution, helping to understand the dataset composition. Figure A3 shows the number of images containing each class, with "Small vehicle" and "Person" being the most prevalent, appearing in 4724 and 3347 images, respectively. The total count of images across all classes is 12,812, which exceeds the total number of unique images (6054). This discrepancy indicates that multiple classes often appear in the same image, resulting in numerous annotations per image.

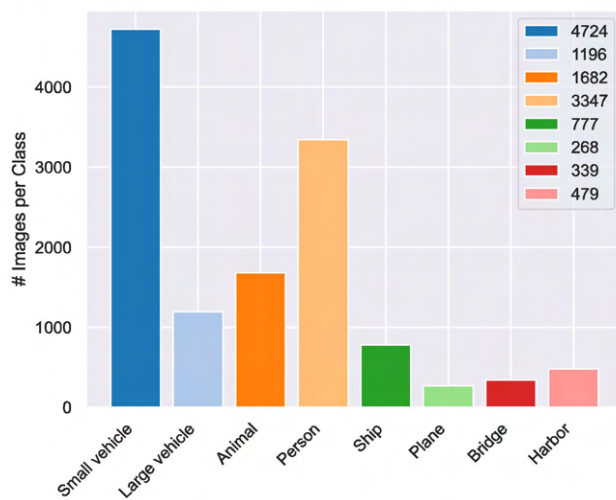

**Figure A3.** Histogram showing the number of images per class in the Common-VALID dataset.

Figure A4 displays the number of instances for each class in the dataset, with "Small vehicle" and "Person" also being the most predominant in terms of instance count. The total number of instances across the 8 different object classes is 70,799.

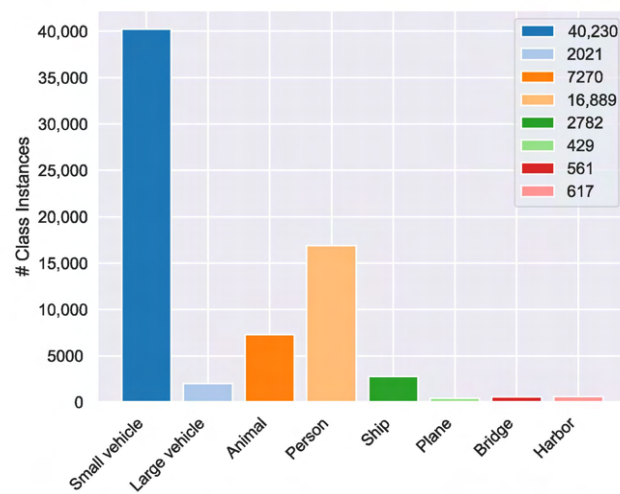

**Figure A4.** Histogram showing the number of instances per class in the Common-VALID dataset.

## Appendix C. Analysis of Failed Freezing Approaches on the Bird's Nest Dataset

This appendix revisits the cases in which layer freezing undermined performance, focusing on the Bird's Nest dataset with YOLOv10. When the FR3 configuration (keeping the first 23 layers fixed) was applied, detection accuracy collapsed. For example, YOLOv10n saw its mAP@50 fall from 0.629 to 0.055 (−91.3%), versus YOLOv10s from 0.633 to 0.192 (−69.7%), YOLOv10m from 0.641 to 0.227 (−64.6%), and YOLOv10l from 0.676 to 0.290 (−57.1%). Notably, this degradation was unique to Bird's Nest. The same FR3 setup remained reasonably effective on the other benchmark datasets (see Table 7).

A discernible connection emerges between the share of frozen parameters and the magnitude of performance loss. Under FR3, only 0.94 M of 2.8 M parameters remain trainable in YOLOv10n (66.5%), while YOLOv10s, m, and l expose 1.69 M/8.1 M (79.1%), 2.37 M/16.6 M (85.7%), and 2.95 M/25.9 M (88.6%) parameters, respectively. Smaller variants not only have fewer total parameters but also expose a much smaller portion of the network to learning. Preliminary observations suggest that models with fewer than approximately 1–1.5 M unfrozen parameters tend to perform worse on the Bird's Nest dataset, which features rich data augmentations.

This trend is tentatively supported by gradient-norm analysis. At the start of training on Bird's Nest, the gradient norm ratio between frozen and fully fine-tuned runs remains between 0.56 and 0.59—about 40–50% lower than reference values—and lacks the characteristic early peaks often associated with escaping saddle points. In contrast, for the InsPLAD-Det dataset (where FR3 performs well), the initial ratio ranges from 0.68 to 0.70 and retains more pronounced, stable gradient dynamics. These patterns are visualized in Figure A5.

Bird's Nest proves vulnerable to aggressive freezing due resulting from three possible interrelated causes: heavy augmentations create distribution shifts incompatible with frozen pre-trained features, fine-grained textures require early-layer adaptation that FR3 prevents, and smaller models lack sufficient trainable capacity to compensate. This combination forces adaptation into an undersized parameter set, causing optimization to stall when the gradient flow drops below critical thresholds. Based on these findings, we recommend limiting parameter freezing to less than 50% when working with augmented or distribution-shifted datasets, while monitoring gradient ratios during early training, as values below 0.6 reliably indicate potential convergence failure. Most importantly, freezing strategies should be tailored to individual dataset characteristics, rather than applying universal configurations across diverse domains.

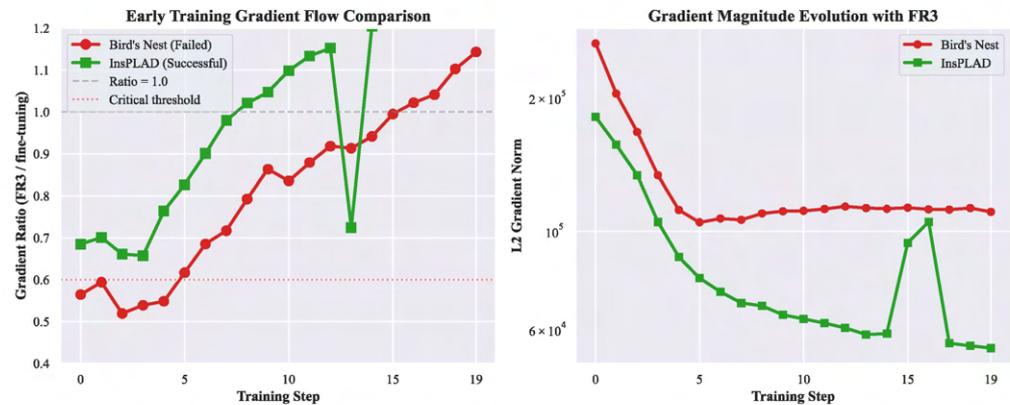

**Figure A5.** Gradient flow comparison between Bird's Nest (failure) and InsPLAD-Det (success). (**Left**) ratio of gradient norms (FR3/fine-tune) over the first 20 steps. (**Right**) absolute gradient magnitudes. The Bird's Nest curves remain below the empirical 0.6 threshold, indicating restricted learning capacity.